\pdfoutput=1

\documentclass[11pt]{article}

\usepackage[]{acl_latex}

\usepackage{times}
\usepackage{latexsym}

\usepackage[T1]{fontenc}

\usepackage[utf8]{inputenc}

\usepackage{graphicx}
\usepackage{multirow}
\usepackage{float}
\usepackage{microtype}

\title{CK-Transformer: Commonsense Knowledge Enhanced Transformers for \\ Referring Expression Comprehension}

\author{Zhi Zhang \\
  ILLC, University of Amsterdam \\ 
  \texttt{z.zhang@uva.nl} \\\And
 Helen Yannakoudakis \\
  Dept. of Informatics, King’s College London\\ 
  \texttt{helen.yannakoudakis@kcl.ac.uk}\\
  \AND
 Xiantong Zhen \\
 United Imaging  \\
  \texttt{zhenxt@gmail.com} \\\And
 Ekaterina Shutova\\
 ILLC, University of Amsterdam  \\
  \texttt{ e.shutova@uva.nl} \\}

\begin{document}
\maketitle
\begin{abstract}
The task of multimodal referring expression comprehension (REC), aiming at localizing an image region described by a natural language expression, has recently received increasing attention within the research comminity. 
In this paper, we specifically focus on referring expression comprehension with commonsense knowledge (KB-Ref), a task which typically requires reasoning beyond spatial, visual or semantic information. 
We propose a novel framework for Commonsense Knowledge Enhanced Transformers (CK-Transformer) which effectively integrates commonsense knowledge into the representations of objects in an image, facilitating identification of the target objects referred to by the expressions. We conduct extensive experiments on several benchmarks for the task of KB-Ref. Our results show that the proposed CK-Transformer achieves a new state of the art, with an absolute improvement of $3.14\%$ accuracy over the existing state of the art \footnote{The code will be available in \url{https://github.com/FightingFighting/CK-Transformer}}.
\end{abstract}

\begin{figure*}
\centerline{\includegraphics[width=0.75\textwidth]{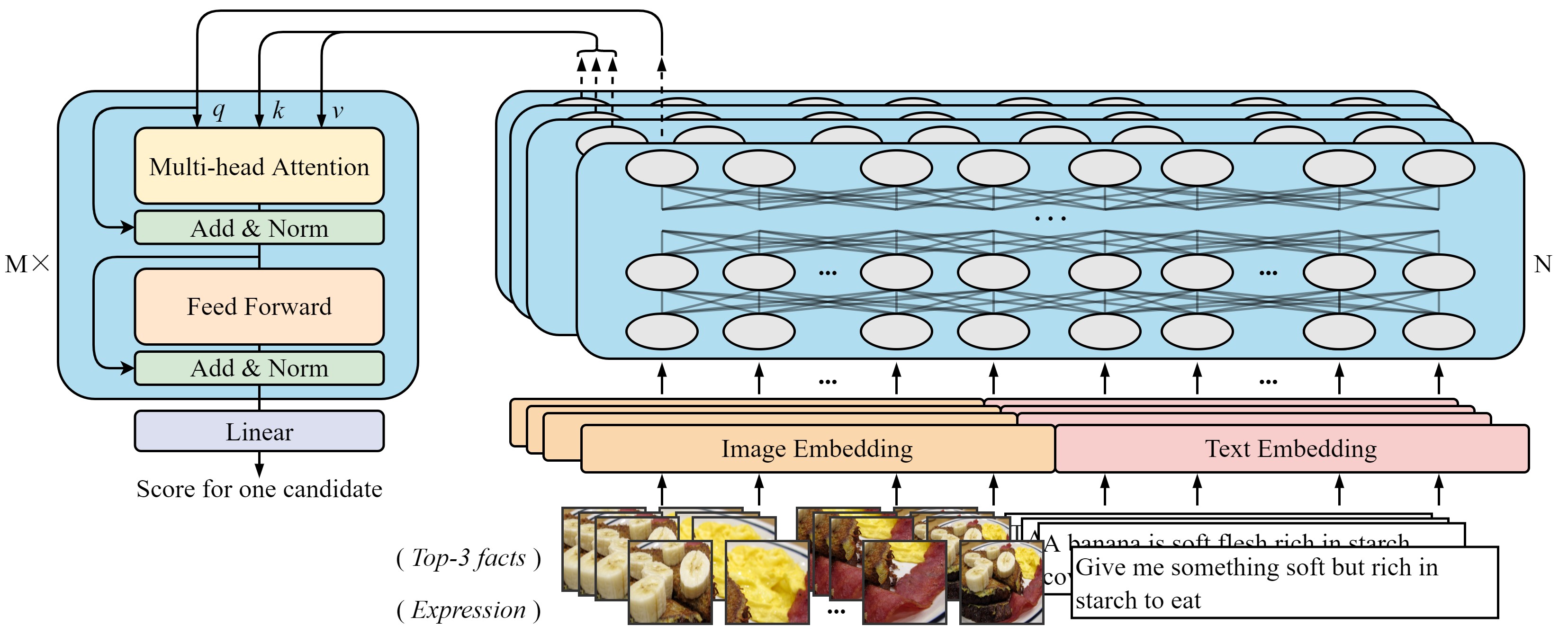}}
\caption{CK-Transformer. For each candidate (the first one in the figure), given an expression, a set of visual region candidates and top-K facts (K$=$3 in the figure), the model first encodes the expression and all top-K facts into corresponding multi-modal features, then fuses these features and maps them into a matching score for the candidate.}
\label{cktransformer}
\end{figure*}

\section{Introduction}
Referring expression comprehension (REC) aims at locating a target object/region in an image given a natural language expression as input. The nature of the task requires multi-modal reasoning and joint visual and language understanding. In the past few years, several REC tasks and datasets have been proposed, such as 
RefCOCO \cite{yu2016modeling}, RefCOCOg \cite{mao2016generation} and RefCOCO+ \cite{yu2016modeling} (RefCOCOs).
These `conventional' REC tasks typically focus on identifying an object based on visual or spatial information of the object, such as its colour, shape, location, etc.; therefore primarily evaluating a model's reasoning abilities over visual attributes and spatial relationships. 

In practice, however, people often describe an object using non-visual or spatial information -- consider, for example, the sentence (expression) ``Give me something soft but rich in starch to eat'' \cite{wang2020give}. Such instances require reasoning beyond spatial and visual attributes, and need to be interpreted with respect to the common sense knowledge (fact) embedded in the expressions, such as knowledge about which kind of objects are edible, soft and rich in starch in the given image.
Recently, \citet{wang2020give} proposed a new dataset, KB-Ref, to evaluate the reasoning ability of a model over not only visual and spatial features but also commonsense knowledge. The dataset is devised such that at least one piece of fact from a knowledge base (KB) is required for a target object (referred to by an expression) to be identified. 



Therefore, searching for appropriate facts from a KB is also crucial part in KB-Ref. In contrast to the only existing work \cite{wang2020give}, in which for each object candidate, the top-K facts with the highest cosine similarity between the averaged Word2Vec \cite{mikolov2013distributed} embedding of the fact and the given expression are maintained, our framework focuses on multi-modal embedding and reasoning simultaneously over both the expression and the image to identify the top-K facts.
Multi-modal features encode richer information helping to improve reasoning over varying (semantic) contexts and identification of relevant facts; for example, the above example of expression can be answered with the object ``banana'' in an image (or, equivalently, with the object ``mushed potato'' in another image). 

In this paper, we propose a novel multi-modal framework for KB-Ref -- Commonsense Knowledge Enhanced Transformers (CK-Transformer, CK-T for short) -- that integrates (top-K) facts into all object candidates in an image for better identification of the target object. 
Specifically, our contributions are four-fold:
1) We propose the CK-T (see Figute \ref{cktransformer}) that effectively integrates diverse input from different modalities: vision, referring expressions and facts; 
2) To the best of our knowledge, our approach is the first that introduces visual information into the identification of (top-K) relevant facts; 
3) Our approach achieves a new state of the art using only top-3 facts per (candidate) object, which is furthermore substantially more efficient compared to existing work utilizing as much as top-50 facts; 4) We introduce facts into `conventional' REC tasks, leading to improved performance.

\section{Related Work}
\paragraph{Referring expression comprehension with commonsense knowledge}

Different from conventional REC tasks (see Appendix \ref{sec:appendix related work REC} for details), KB-Ref focuses on querying objects given an expression that requires commonsense knowledge reasoning. The authors benchmarked a baseline model, ECIFA, for integration of facts, expression and image, and selects the target object by comparing the match scores between the image features and corresponding top-K fact features for all object candidates in the image \cite{wang2020give}. In our framework, we select top-K facts for each candidate by comparing the cosine similarity between the fact and expression embedding, where the embeddings are generated from a multi-modal encoder rather than a text encoder used in the ECIFA model.


\paragraph{Pre-trained vision--language encoders} 
Several pre-trained multi-modal encoders \cite{su2019vl, li2019visualbert, chen2020uniter, tan2019lxmert} have been proposed, achieving state-of-the-art results on  vision--language tasks.
Currently, UNITER \cite{chen2020uniter} as one of powerful pre-trained encoders achieves the best performance on REC tasks (RefCOCOs). In this paper, we adapt UNITER such that it is used as a multi-modal encoder in fact search and as part of the CK-T.


\section{Methodology}
We formulate KB-Ref as a classification problem based on an image $I$ consisting of a set of candidates (image regions) $I=\{c_j\}_{j=1}^n$ obtained from either ground-truth labels or predictions of a pre-trained object detector. Specifically, given an expression $e$, an image $I$ and a KB, we first search for top-K facts $F_i^K=\{f_j\}_{j=1}^k$ from the KB for each candidate $c_i$, and then feed $e$, $I$, and $\textit{F}_I^K$ (the selected facts over $I$) into our CK-T simultaneously to predict the target object over all candidates.

\subsection{Image-based fact search}
For each candidate $c_i$ in a given image, we retrieve all the facts from the KB (see Appendix \ref{sec:appendix Dataset} for details on the KB used in our framework) according to its category (e.g., a candidate object may belong to category `car'). Then, we calculate the cosine similarity between the facts and the given expression, where the similarities are obtained from a similarity extractor which we train by adapting UNITER. Specifically, given image--expression and image--fact pairs as input, we extract expression and fact features respectively from the position of the cross-modality output of UNITER (corresponding to the input of [CLS] token, see Appendix \ref{sec:appendix UNITER} for details), and then calculate the cosine similarity between the two. During training, inspired by \citet{devlin2018bert}, we replace $50\%$ of ground-truth facts with random facts from the KB (with a similarity of $0$), to help the model better distinguish useful facts from non-useful ones. 
Finally, we maintain top-K facts $F_i^K$ with higher similarities to the expression for each candidate $c_i$. 

\subsection{Commonsense Knowledge Enhanced Transformer} The CK-T consists of a bi-modal encoder (see \ref{sec:bi-modal encoder}) and a fact-aware classifier (see \ref{sec:Fact-aware classifier}). 

\subsubsection{Bi-modal encoder}
\label{sec:bi-modal encoder}
The bi-modal encoder (initialized by UNITER-base with N=12 layers \cite{chen2020uniter}) integrates two modalities: image and text ($e$ or $f_i$). Specifically, after generating the input embedding $\textit{\textbf{E}}_{Inp}$ consisting of image and text embedding (same with UNITER, see Appendix \ref{sec:appendix Image embedding} for details), 
for each candidate $c_i$, we extract the expression-aware and fact-aware object features respectively ($\textit{\textbf{f}}_i^e$ and  $\textit{\textbf{f}}_i^f$) from the position of the visual output corresponding to $c_i$ in the same encoder, based on the input of all candidates $I$, and $e$ or $f_i$. 


\subsubsection{Fact-aware classifier}
\label{sec:Fact-aware classifier}
The fact-aware classifier is composed of multi-head attention layers and fully connected layers. For each candidate $c_i$, $\textit{\textbf{f}}_i^e$ and $\textit{\textbf{F}}_i^f$ (all K fact-aware object features for $c_i$) are fed into the integrator simultaneously ($\textit{Key}$ and $\textit{Value}$ are from $\textit{\textbf{F}}_i^f$, and $\textit{Query}$ is from $\textit{\textbf{f}}_i^e$), and fused into one three-source object features $\textit{\textbf{f}}_i^t$ (image, expression and top-K facts). 

Finally, $\textit{\textbf{f}}_i^t$ is mapped into a match score $s_i$ for $c_i$ by a linear layer, and the optimization objective is to minimize the cross-entropy loss over all scores $\{s_j\}_{j=1}^n$ corresponding to all candidates $I$.

\section{Results}
We compare our CK-T to existing approaches on KB-Ref task without and with facts. Then we explore the importance of introducing visual information for fact search. Furthermore, we introduce facts into the traditional RefCOCOs dataset, which was collected from MSCOCO \cite{lin2014microsoft} but differs in the types of expressions and object candidate settings. 
We extract image region features using an off-the-shelf detector, Faster R-CNN with ResNet-101 \cite{ren2015faster}, based on bounding boxes (bbxes) (ground-truth labels or predicted results from the detector). See Appendix \ref{sec:appendix Dataset} and \ref{sec:appendix Experimental settings} for details about these datasets and experiment setting\footnote{Including the efficiency discussion about our model}. Through parameter search on K and M (see Figure \ref{Accuracy on different top-K facts} and \ref{Accuracy on different integrator layer} in Appendix \ref{sec:appendix parameter search}), we keep M = 2 Fact-aware classifier blocks and top-3 facts for each candidate.



\paragraph{Ground-truth bounding boxes and categories}
\begin{table}
\centering
\begin{tabular}{lcc}
\hline 
\multicolumn{1}{c}{\multirow{2}{*}{\textbf{Model}}} & \multicolumn{2}{c}{\textbf{Accuracy (\%)}}  \\
 & \textbf{Val} & \textbf{Test} \\ 
\hline
CMN \cite{hu2017modeling} & 41.28 & 40.03\\
SLR \cite{yu2017joint} & 44.03 & 42.92\\
VC \cite{niu2019variational} & 44.63 & 43.59\\
LGARNs \cite{wang2019neighbourhood} & 45.11 & 44.27\\
MAttNet \cite{yu2018mattnet} & 46.86 & 46.03\\
ECIFA-nf \cite{wang2020give} & 37.95 & 35.16\\
CK-T-nf (Ours) & \textbf{58.02} & \textbf{57.53} \\ 
\hline
ECIFA \cite{wang2020give} & 59.45 & 58.97 \\
MAtt+E \cite{wang2020give} & 64.08 & 63.57\\

CK-T-Word2Vec & 60.40 & 61.39\\
CK-T-Uw/oImage  & 64.44 & 64.78 \\

CK-T (Ours) & \textbf{65.62} & \textbf{66.71} \\
\hline
Human & \multicolumn{1}{c}{$-$} & 90.31 \\
\hline
\end{tabular}
\caption{\label{KB-Ref with gt bbxes} Accuracy on KB-Ref dataset without and with facts (top and bottom part, respectively) using ground-truth bounding boxes and object categories.}
\end{table}
By following \citet{wang2020give}, we report our results on KB-Ref without and with facts. As can be seen in Table \ref{KB-Ref with gt bbxes} (top), CK-T-nf, a version of CK-T without facts\footnote{all word tokens in fact sentences are replaced with only one [MASK] token.}, achieves an accuracy of $57.53\%$ on the test set, outperforming existing approaches that do not utilize facts by approximately $11\%-22\%$. 
At the bottom part of the table we can see that our fact-enhanced CK-T model achieves the highest accuracy ($66.71\%$) on the test set, which is $7.74\%$ higher than that of ECIFA (a baseline model proposed by \citet{wang2020give}), and $3.14\%$ higher than MAtt+E\footnote{\citet{wang2020give} introduces their facts fusion module --Episodic Memory Module (E)--into MAttNet model (Matt) \cite{yu2018mattnet} widely used for conventional REC.}. It is worth noting that both ECIFA and MAtt+E incorporate the top-50 facts for each candidate, which is considerably higher compared to top-3 facts in our CK-T. We surmise this is due to the fact that our fact search approach utilizes multi-modal fact and expression embeddings. 

\paragraph{Predicted bounding boxes and categories}
To facilitate a fair comparison with ECIFA-d \cite{wang2020give}, we also use the maximum 10 detected bbxes for each image (CK-T-m10). As can be seen in Table \ref{KB-Ref with detected bbxes}, CK-T-m10 achieves an accuracy which is $\approx5\%$ higher than that of ECIFA-d on the test set.
CK-T-m100, a variant using at most 100 detected bbxes achieves a substantial improvement with $\approx7\%$, compared with CK-T-m10.
This difference is primarily due to the increase in the number of correctly detected bbxes and predicted categories. Specifically, we find that with the top-100 bbxes, the number of samples containing the target bbxes rises from $18,901$ to $31,653$, while among these target bbxes, the number of correctly predicted categories grows from $11,324$ to $15,928$, out of a total of $43,284$ 
samples in the KB-Ref dataset. This can also explain the dramatic decline on the accuracy between CK-T and CK-T-m10.
\begin{table}
\centering
\begin{tabular}{lcc}
\hline 
\multicolumn{1}{c}{\multirow{2}{*}{\textbf{Model}}} & \multicolumn{2}{c}{\textbf{Accuracy (\%)}}  \\
 & \textbf{Val} & \textbf{Test} \\ 
 \hline
ECIFA-d \cite{wang2020give} & 24.11 & 23.82 \\
CK-T-m10 (Ours) & 28.33 & 28.71 \\
CK-T-m100 (Ours) & \textbf{35.66} & \textbf{35.96} \\
\hline
\end{tabular}
\caption{\label{KB-Ref with detected bbxes} Accuracy on KB-Ref using predicted bbxes and object categories.}
\end{table}

\paragraph{Incorporating image features into fact search}
\label{sec: search method}

We experiment with various approaches to fact search and evaluate their effectiveness on KB-Ref (Table \ref{KB-Ref with gt bbxes}). 
We first utilize top-k facts searched in \cite{wang2020give}, where they use a pre-trained Word2Vec (Skip-Gram) \cite{mikolov2013distributed}
for searching facts (CK-T-Word2Vec).
Then, we also selected facts from similarity predictors based on only text as input (CK-T-Uw/oImage)\footnote{Inspired by \citet{frank2021vision}, we replace all object candidate feature with the average of all image region features.}, instead of image-text pairs in CK-T. As shown in Table \ref{KB-Ref with gt bbxes}, both CK-T-Uw/oImage and CK-T achieve better accuracy on the test set compared to CK-T-Word2Vec. Compared to CK-T-Uw/oImage, CK-T achieves around $2\%$ higher accuracy. This is primarily due to the additional visual information used during fact search (see Appendix \ref{sec:appendix searched fact} for the examples of the selected facts by these fact search methods).

\paragraph{Introducing facts in traditional REC tasks}

\begin{table}
\centering
\begin{tabular}{llcc}
\hline 
\multicolumn{2}{c}{\multirow{2}{*}{\textbf{Task}}} & \multicolumn{2}{c}{\textbf{Accuracy (\%)}}  \\
& & $\textbf{U}_{REC}$ & \textbf{Intro Facts} \\
\hline
\multirow{3}{*}{\begin{tabular}[c]{@{}l@{}}Ref-\\ COCO\end{tabular}}  
& Val & 90.98 & \textbf{91.43}\\
& Test A & 91.50 & \textbf{92.09}\\
& Test B & 90.89 & \textbf{90.95}\\\hline
\multirow{3}{*}{\begin{tabular}[c]{@{}l@{}}Ref-\\ COCO+\end{tabular}} 
& Val & 83.23 & \textbf{83.45} \\
& Test A & 85.09 & \textbf{85.49}\\
& Test B & \textbf{79.08} & \textbf{79.08}\\\hline
\multirow{2}{*}{\begin{tabular}[c]{@{}l@{}}Ref-\\ COCOg\end{tabular}} 
& Val & 86.23 & \textbf{87.21}\\
& Test & 85.79 & \textbf{87.59} \\\hline
\end{tabular}
\caption{\label{Introducing commonsense} Introducing facts into RefCOCO, RefCOCO+ and RefCOCOg. RefCOCO and RefCOCO+ have two different test sets, Test A and Test B, containing multiple persons and multiple objects in images respectively.}
\end{table}
We incorporate facts from the KB used in KB-Ref into the tasks of RefCOCOs using CK-T. Table \ref{Introducing commonsense} shows the results comparison based on the ground-truth bbxes and categories (discussion about the detected results can be seen in Appendix \ref{sec:appendix intro facts into refcoco using detected bbxes}). Compared with U$_{REC}$\footnote{\citet{chen2020uniter} achieve state-of-the-art results on RefCOCOs by finetuning UNITER. (We re-finetune the model for fair comparison and conduct McNemar Test)}, the model introducing facts achieves better or equal accuracy on all RefCOCOs tasks, where RefCOCOg is improved more than RefCOCO and RefCOCO+. This is because RefCOCOg has less same-category object candidates in an image compared to RefCOCO and RefCOCO+ (an average of 1.63 and 3.9 per image, respectively) \cite{yu2016modeling}, and thus the retrieved facts integrated into different candidates are diversified (we first retrieve facts using the category), which contributes to distinguishing between candidates. This difference can also be proved in McNemar Test, where we find the change in the proportion of errors is statistically significant after introducing facts as compared to before on RefCOCOg (\textit{p}-value $=$ 1.19$e-$08 $<$ $\alpha=$ 0.05), while the similar proportions are found on RefCOCO and RefCOCO+ (see Appendix \ref{sec:McNemar Test} for details about McNemar Test). The overall impact of commonsense knowledge in traditional REC is, however, not substantial. This is primarily due to much smaller number (78) of categories among the candidates in RefCOCOs, compared to 1805 in the KB-Ref \cite{wang2020give}. This limits the variety of selected facts, therefore impacting the extent to which commonsense knowledge is useful.

\section{Analysis}
\label{sec: appendix Fine-grained analysis}
To investigate in what cases commonsense knowledge helps, we conduct a fine-grained analysis of model performance on the test set of KB-Ref. Specifically, we compare the samples predicted by model with and without facts (CK-T and CK-T-nf) on three aspects: object categories, spatial relationships and the size of the bounding box.

\begin{figure}
    \centering
    \subfigure[Top 10 categories showing most improvement after introducing  facts.]{
        \includegraphics[width=0.47\textwidth]{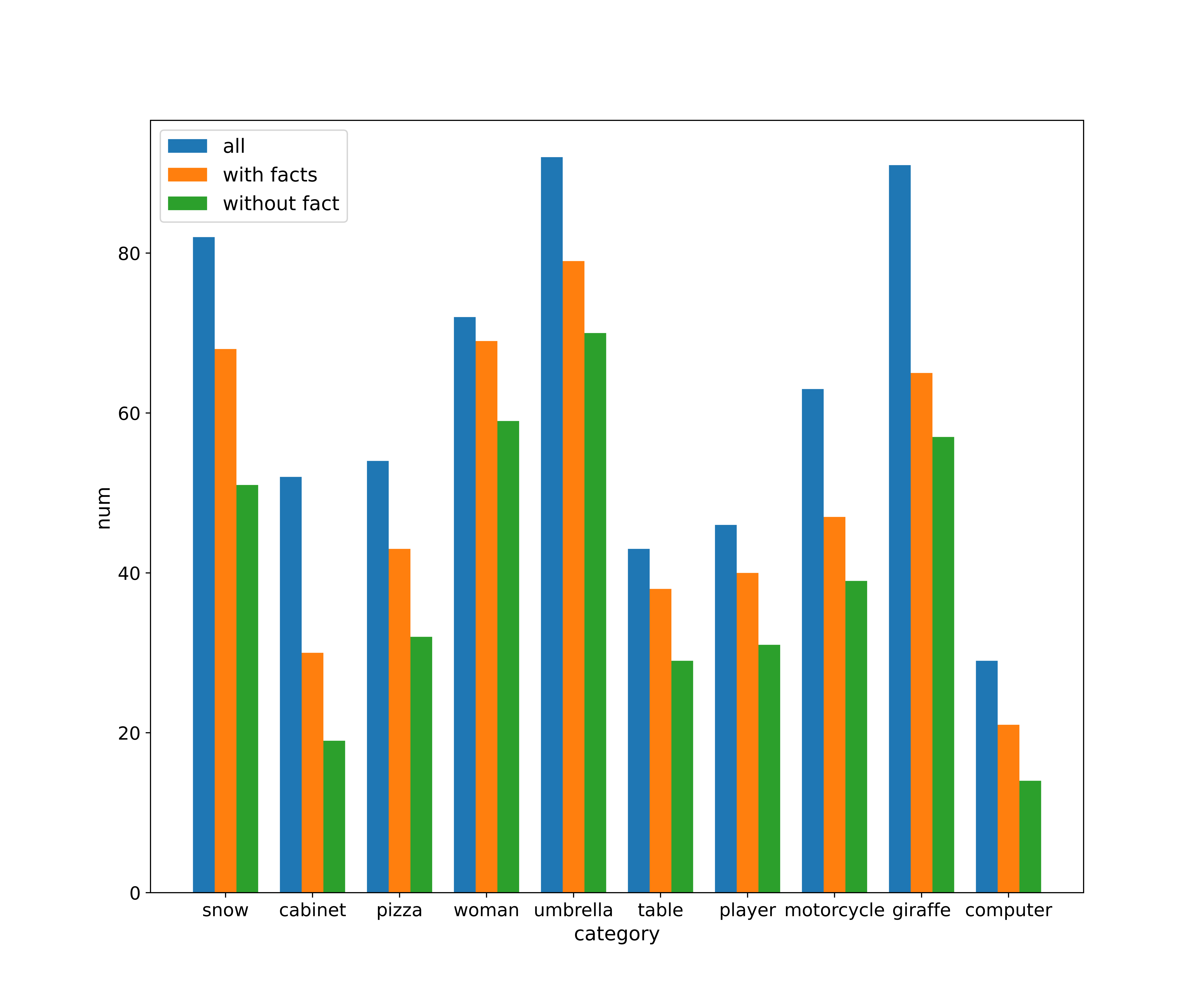}
        \label{fine-grained analysis top 10 categories}
         }

    \centering
    \subfigure[The analysis of spatial relationships.]{
        \includegraphics[width=0.47\textwidth]{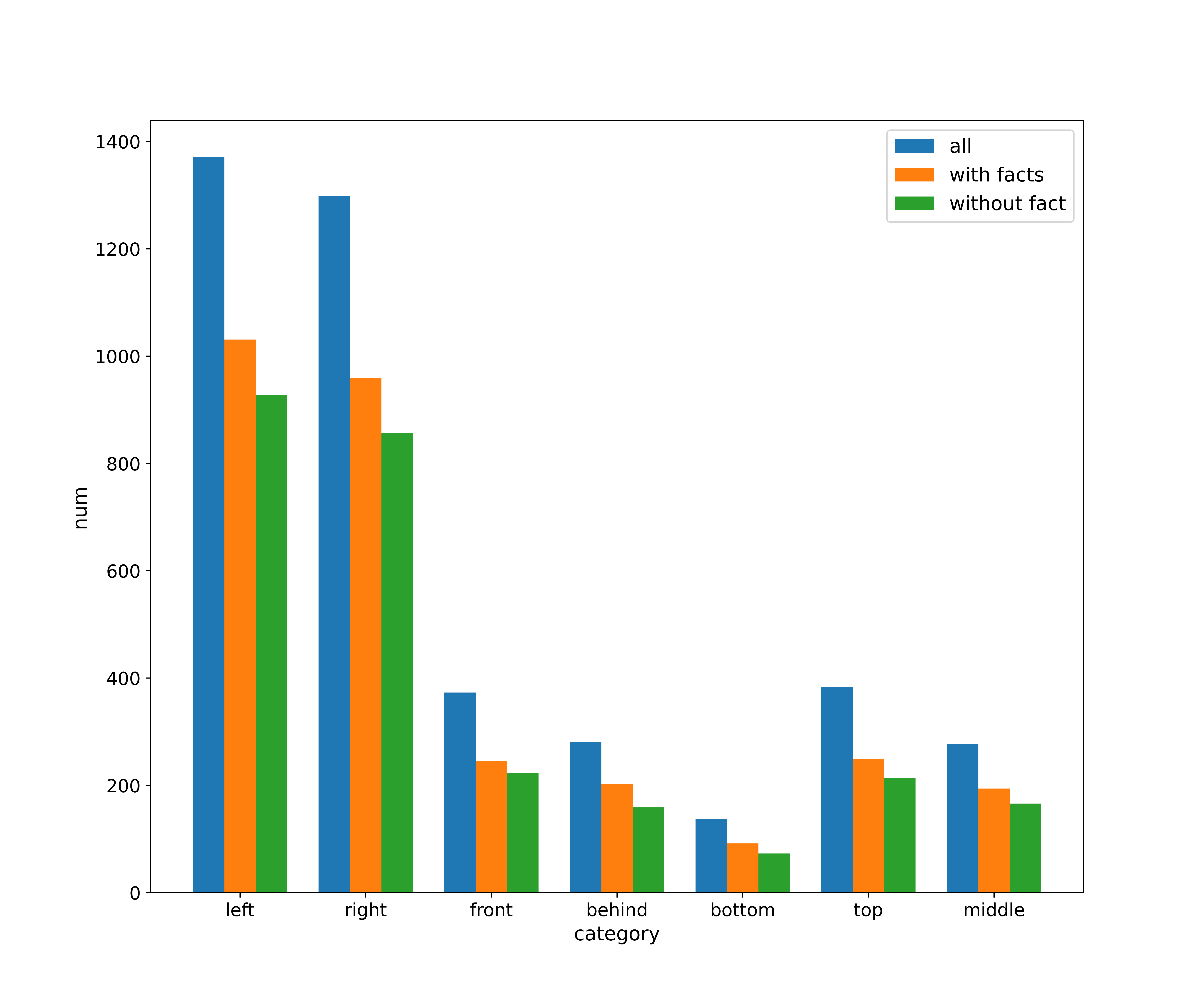}
        \label{fine-grained analysis spatial}
         }

    \centering
    \subfigure[The analysis of different bounding box sizes.]{
        \includegraphics[width=0.47\textwidth]{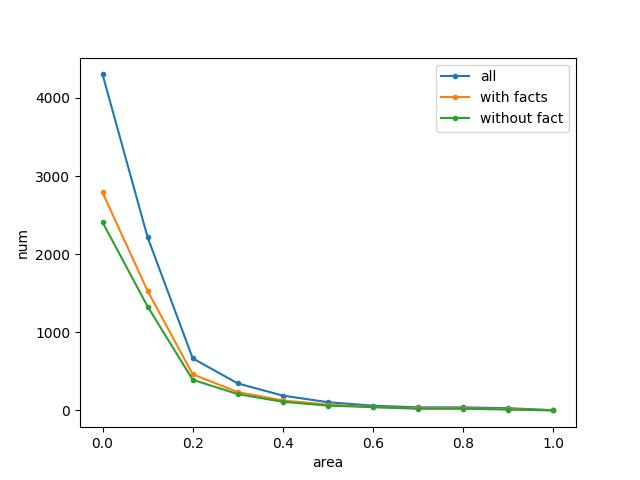}
        \label{fine-grained analysis bbox}
         }
    \caption{Fine-grained analysis. \textit{all}: the total number of samples in the test set; \textit{with fact}: the number of test samples that CK-T predicts correctly; \textit{without fact}: the number of test samples that CK-T-nf predicts correctly.}
    \label{Figure: fine-grained analysis for three aspects}
\end{figure}


\paragraph{Object categories}
The test set contains 1502 categories and CK-T outperforms CK-T-nf on 1347 categories. Top 10 categories for which most improvement is observed are shown in Figure \ref{fine-grained analysis top 10 categories}. In case of the 155 categories that do not show improvement, we find that the average number of samples per category is 6.68, making the results less reliable.


\paragraph{Spatial relationships} We then investigate to what extent solving the REC task with and without facts relies on spatial reasoning, and whether there are particular spatial relationships between objects for which the use of facts is most crucial. Similar to the works of  \cite{kazemzadeh2014referitgame,johnson2017clevr}, we focus on the following spatial relationships: \textit{left, right, front, behind, bottom, top, middle}. As shown in Figure \ref{fine-grained analysis spatial}, the model with facts (CK-T) outperforms that without facts (CK-T-nf) on all spatial relationships.


\paragraph{The size of the bounding box} We then investigate the role of facts when identifying objects of different sizes, using the size of their bounding box as a proxy. 
We use the normalized area of the bounding box as the metric of bbxes size. As shown in Figure \ref{fine-grained analysis bbox}, the facts improve model performance on all bounding box sizes.

\section{Conclusion}
In this paper, we proposed CK-Transformer, which effectively integrates commonsense knowledge and the expression into the representations of the corresponding visual objects for multi-modal reasoning on KB-Ref. Our CK-Transformer achieves a new state-of-the-art performance on KB-Ref using only top-3 most relevant facts. We also demonstrated that visual information is beneficial for fact search. Finally, we show that commonsense knowledge improves conventional REC tasks across three different datasets. 

\section{Limitations}
The computational requirements of our model are affected by the number of facts. Specifically, we train our CK-Transformer for 10000 steps with a batch size of $64$ on one Titan RTX GPU, which takes 2.5, 3, 3.5, 7 days with the number of facts: 3, 5, 10, 20 respectively. The CK-Transformer processes 3.8, 2.8, 2.1, 0.7 samples on average per second at training time and 8.3, 7.3, 6.6, 1.1 samples per second at test time, with these amounts of facts. The computational requirements of our models are thus substantial, and future work should consider improving computational efficiency and thus reducing environmental impact.  

\bibliography{anthology,custom}

\begin{thebibliography}{29}
\expandafter\ifx\csname natexlab\endcsname\relax\def\natexlab#1{#1}\fi

\bibitem[{Anderson et~al.(2018)Anderson, He, Buehler, Teney, Johnson, Gould,
  and Zhang}]{anderson2018bottom}
Peter Anderson, Xiaodong He, Chris Buehler, Damien Teney, Mark Johnson, Stephen
  Gould, and Lei Zhang. 2018.
\newblock Bottom-up and top-down attention for image captioning and visual
  question answering.
\newblock In \emph{Proceedings of the IEEE conference on computer vision and
  pattern recognition}, pages 6077--6086.

\bibitem[{Chen et~al.(2020)Chen, Li, Yu, El~Kholy, Ahmed, Gan, Cheng, and
  Liu}]{chen2020uniter}
Yen-Chun Chen, Linjie Li, Licheng Yu, Ahmed El~Kholy, Faisal Ahmed, Zhe Gan,
  Yu~Cheng, and Jingjing Liu. 2020.
\newblock Uniter: Universal image-text representation learning.
\newblock In \emph{European conference on computer vision}, pages 104--120.
  Springer.

\bibitem[{Devlin et~al.(2018)Devlin, Chang, Lee, and
  Toutanova}]{devlin2018bert}
Jacob Devlin, Ming-Wei Chang, Kenton Lee, and Kristina Toutanova. 2018.
\newblock Bert: Pre-training of deep bidirectional transformers for language
  understanding.
\newblock \emph{arXiv preprint arXiv:1810.04805}.

\bibitem[{Frank et~al.(2021)Frank, Bugliarello, and Elliott}]{frank2021vision}
Stella Frank, Emanuele Bugliarello, and Desmond Elliott. 2021.
\newblock Vision-and-language or vision-for-language? on cross-modal influence
  in multimodal transformers.
\newblock \emph{arXiv preprint arXiv:2109.04448}.

\bibitem[{Hu et~al.(2017)Hu, Rohrbach, Andreas, Darrell, and
  Saenko}]{hu2017modeling}
Ronghang Hu, Marcus Rohrbach, Jacob Andreas, Trevor Darrell, and Kate Saenko.
  2017.
\newblock Modeling relationships in referential expressions with compositional
  modular networks.
\newblock In \emph{Proceedings of the IEEE Conference on Computer Vision and
  Pattern Recognition}, pages 1115--1124.

\bibitem[{Hu et~al.(2016)Hu, Xu, Rohrbach, Feng, Saenko, and
  Darrell}]{hu2016natural}
Ronghang Hu, Huazhe Xu, Marcus Rohrbach, Jiashi Feng, Kate Saenko, and Trevor
  Darrell. 2016.
\newblock Natural language object retrieval.
\newblock In \emph{Proceedings of the IEEE Conference on Computer Vision and
  Pattern Recognition}, pages 4555--4564.

\bibitem[{Johnson et~al.(2017)Johnson, Hariharan, Van Der~Maaten, Fei-Fei,
  Lawrence~Zitnick, and Girshick}]{johnson2017clevr}
Justin Johnson, Bharath Hariharan, Laurens Van Der~Maaten, Li~Fei-Fei,
  C~Lawrence~Zitnick, and Ross Girshick. 2017.
\newblock Clevr: A diagnostic dataset for compositional language and elementary
  visual reasoning.
\newblock In \emph{Proceedings of the IEEE conference on computer vision and
  pattern recognition}, pages 2901--2910.

\bibitem[{Kazemzadeh et~al.(2014)Kazemzadeh, Ordonez, Matten, and
  Berg}]{kazemzadeh2014referitgame}
Sahar Kazemzadeh, Vicente Ordonez, Mark Matten, and Tamara Berg. 2014.
\newblock Referitgame: Referring to objects in photographs of natural scenes.
\newblock In \emph{Proceedings of the 2014 conference on empirical methods in
  natural language processing (EMNLP)}, pages 787--798.

\bibitem[{Krishna et~al.(2016)Krishna, Zhu, Groth, Johnson, Hata, Kravitz,
  Chen, Kalantidis, Li, Shamma et~al.}]{krishna2016visual}
Ranjay Krishna, Yuke Zhu, Oliver Groth, Justin Johnson, Kenji Hata, Joshua
  Kravitz, Stephanie Chen, Yannis Kalantidis, Li-Jia Li, David~A Shamma, et~al.
  2016.
\newblock Visual genome: Connecting language and vision using crowdsourced
  dense image annotations.
\newblock \emph{arXiv preprint arXiv:1602.07332}.

\bibitem[{Li et~al.(2019)Li, Yatskar, Yin, Hsieh, and Chang}]{li2019visualbert}
Liunian~Harold Li, Mark Yatskar, Da~Yin, Cho-Jui Hsieh, and Kai-Wei Chang.
  2019.
\newblock Visualbert: A simple and performant baseline for vision and language.
\newblock \emph{arXiv preprint arXiv:1908.03557}.

\bibitem[{Lin et~al.(2014)Lin, Maire, Belongie, Hays, Perona, Ramanan,
  Doll{\'a}r, and Zitnick}]{lin2014microsoft}
Tsung-Yi Lin, Michael Maire, Serge Belongie, James Hays, Pietro Perona, Deva
  Ramanan, Piotr Doll{\'a}r, and C~Lawrence Zitnick. 2014.
\newblock Microsoft coco: Common objects in context.
\newblock In \emph{European conference on computer vision}, pages 740--755.
  Springer.

\bibitem[{Loshchilov and Hutter(2017)}]{loshchilov2017decoupled}
Ilya Loshchilov and Frank Hutter. 2017.
\newblock Decoupled weight decay regularization.
\newblock \emph{arXiv preprint arXiv:1711.05101}.

\bibitem[{Mao et~al.(2016)Mao, Huang, Toshev, Camburu, Yuille, and
  Murphy}]{mao2016generation}
Junhua Mao, Jonathan Huang, Alexander Toshev, Oana Camburu, Alan~L Yuille, and
  Kevin Murphy. 2016.
\newblock Generation and comprehension of unambiguous object descriptions.
\newblock In \emph{Proceedings of the IEEE conference on computer vision and
  pattern recognition}, pages 11--20.

\bibitem[{Mikolov et~al.(2013)Mikolov, Sutskever, Chen, Corrado, and
  Dean}]{mikolov2013distributed}
Tomas Mikolov, Ilya Sutskever, Kai Chen, Greg~S Corrado, and Jeff Dean. 2013.
\newblock Distributed representations of words and phrases and their
  compositionality.
\newblock In \emph{Advances in neural information processing systems}, pages
  3111--3119.

\bibitem[{Niu et~al.(2019)Niu, Zhang, Lu, and Chang}]{niu2019variational}
Yulei Niu, Hanwang Zhang, Zhiwu Lu, and Shih-Fu Chang. 2019.
\newblock Variational context: Exploiting visual and textual context for
  grounding referring expressions.
\newblock \emph{IEEE transactions on pattern analysis and machine
  intelligence}, 43(1):347--359.

\bibitem[{Ordonez et~al.(2011)Ordonez, Kulkarni, and Berg}]{ordonez2011im2text}
Vicente Ordonez, Girish Kulkarni, and Tamara Berg. 2011.
\newblock Im2text: Describing images using 1 million captioned photographs.
\newblock \emph{Advances in neural information processing systems},
  24:1143--1151.

\bibitem[{Ren et~al.(2015)Ren, He, Girshick, and Sun}]{ren2015faster}
Shaoqing Ren, Kaiming He, Ross Girshick, and Jian Sun. 2015.
\newblock Faster r-cnn: Towards real-time object detection with region proposal
  networks.
\newblock \emph{Advances in neural information processing systems}, 28:91--99.

\bibitem[{Sharma et~al.(2018)Sharma, Ding, Goodman, and
  Soricut}]{sharma2018conceptual}
Piyush Sharma, Nan Ding, Sebastian Goodman, and Radu Soricut. 2018.
\newblock Conceptual captions: A cleaned, hypernymed, image alt-text dataset
  for automatic image captioning.
\newblock In \emph{Proceedings of the 56th Annual Meeting of the Association
  for Computational Linguistics (Volume 1: Long Papers)}, pages 2556--2565.

\bibitem[{Speer et~al.(2017)Speer, Chin, and Havasi}]{speer2017conceptnet}
Robyn Speer, Joshua Chin, and Catherine Havasi. 2017.
\newblock Conceptnet 5.5: An open multilingual graph of general knowledge.
\newblock In \emph{Thirty-first AAAI conference on artificial intelligence}.

\bibitem[{Su et~al.(2019)Su, Zhu, Cao, Li, Lu, Wei, and Dai}]{su2019vl}
Weijie Su, Xizhou Zhu, Yue Cao, Bin Li, Lewei Lu, Furu Wei, and Jifeng Dai.
  2019.
\newblock Vl-bert: Pre-training of generic visual-linguistic representations.
\newblock \emph{arXiv preprint arXiv:1908.08530}.

\bibitem[{Tan and Bansal(2019)}]{tan2019lxmert}
Hao Tan and Mohit Bansal. 2019.
\newblock Lxmert: Learning cross-modality encoder representations from
  transformers.
\newblock \emph{arXiv preprint arXiv:1908.07490}.

\bibitem[{Tandon et~al.(2017)Tandon, De~Melo, and Weikum}]{tandon2017webchild}
Niket Tandon, Gerard De~Melo, and Gerhard Weikum. 2017.
\newblock Webchild 2.0: Fine-grained commonsense knowledge distillation.
\newblock In \emph{Proceedings of ACL 2017, System Demonstrations}, pages
  115--120.

\bibitem[{Wang et~al.(2020)Wang, Liu, Li, and Wu}]{wang2020give}
Peng Wang, Dongyang Liu, Hui Li, and Qi~Wu. 2020.
\newblock Give me something to eat: Referring expression comprehension with
  commonsense knowledge.
\newblock In \emph{Proceedings of the 28th ACM International Conference on
  Multimedia}, pages 28--36.

\bibitem[{Wang et~al.(2019)Wang, Wu, Cao, Shen, Gao, and
  Hengel}]{wang2019neighbourhood}
Peng Wang, Qi~Wu, Jiewei Cao, Chunhua Shen, Lianli Gao, and Anton van~den
  Hengel. 2019.
\newblock Neighbourhood watch: Referring expression comprehension via
  language-guided graph attention networks.
\newblock In \emph{Proceedings of the IEEE/CVF Conference on Computer Vision
  and Pattern Recognition}, pages 1960--1968.

\bibitem[{Wu et~al.(2016)Wu, Schuster, Chen, Le, Norouzi, Macherey, Krikun,
  Cao, Gao, Macherey et~al.}]{wu2016google}
Yonghui Wu, Mike Schuster, Zhifeng Chen, Quoc~V Le, Mohammad Norouzi, Wolfgang
  Macherey, Maxim Krikun, Yuan Cao, Qin Gao, Klaus Macherey, et~al. 2016.
\newblock Google's neural machine translation system: Bridging the gap between
  human and machine translation.
\newblock \emph{arXiv preprint arXiv:1609.08144}.

\bibitem[{Yu et~al.(2018)Yu, Lin, Shen, Yang, Lu, Bansal, and
  Berg}]{yu2018mattnet}
Licheng Yu, Zhe Lin, Xiaohui Shen, Jimei Yang, Xin Lu, Mohit Bansal, and
  Tamara~L Berg. 2018.
\newblock Mattnet: Modular attention network for referring expression
  comprehension.
\newblock In \emph{Proceedings of the IEEE Conference on Computer Vision and
  Pattern Recognition}, pages 1307--1315.

\bibitem[{Yu et~al.(2016)Yu, Poirson, Yang, Berg, and Berg}]{yu2016modeling}
Licheng Yu, Patrick Poirson, Shan Yang, Alexander~C Berg, and Tamara~L Berg.
  2016.
\newblock Modeling context in referring expressions.
\newblock In \emph{European Conference on Computer Vision}, pages 69--85.
  Springer.

\bibitem[{Yu et~al.(2017)Yu, Tan, Bansal, and Berg}]{yu2017joint}
Licheng Yu, Hao Tan, Mohit Bansal, and Tamara~L Berg. 2017.
\newblock A joint speaker-listener-reinforcer model for referring expressions.
\newblock In \emph{Proceedings of the IEEE Conference on Computer Vision and
  Pattern Recognition}, pages 7282--7290.

\bibitem[{Zhang et~al.(2018)Zhang, Niu, and Chang}]{zhang2018grounding}
Hanwang Zhang, Yulei Niu, and Shih-Fu Chang. 2018.
\newblock Grounding referring expressions in images by variational context.
\newblock In \emph{Proceedings of the IEEE Conference on Computer Vision and
  Pattern Recognition}, pages 4158--4166.

\end{thebibliography}
\bibliographystyle{acl_natbib}

\clearpage
\appendix

\section{Referring expression comprehension}
\label{sec:appendix related work REC}
Early approaches to REC use
joint embedding of image and language by combination of Convolutional Neural Networks (CNNs) and Recurrent Neural Networks (RNNs), and predict the target object that has the maximum probability given an input expression and an image \cite{mao2016generation, hu2016natural, zhang2018grounding}. In order to model different types of information encoded in input expression (subject appearance, location, and relationship to other objects), subsequent work used modular (attention) networks, to ``match'' the input to corresponding regions in the image, predicting as the target the region with the highest matched score \cite{hu2017modeling, yu2018mattnet}. 

\section{UNITER}
\label{sec:appendix UNITER}
UNITER is trained using four pre-training tasks, Masked Language Modeling (MLM), Masked Region Modeling (MRM), Image--Text Matching (ITM), and Word--Region Alignment (WRA), on four large-scale image--text datasets, COCO \cite{lin2014microsoft}, Visual Genome \cite{krishna2016visual}, Conceptual Captions \cite{sharma2018conceptual}, and SBU Captions \cite{ordonez2011im2text}. This enables UNITER to capture  fine-grained alignments between images and language. The architecture of UNITER is similar to BERT \cite{devlin2018bert} apart from the input and the output. Specifically, the input consists of an image (a set of visual region candidates), a sentence and [CLS] token, and they respectively lead to different outputs, i.e. vision output, language output and cross-modality output on the top of UNITER.

\section{Input embedding}
Same with UNITER, we extract the input embeddings $\textit{\textbf{E}}_{Inp}$ consisting of an image and a text embedding corresponding to the object candidate $I$ and text (an expression $e$ or a fact $f_i$) respectively.

\label{sec:appendix Image embedding}
\paragraph{Image embedding} The image embedding $\textit{\textbf{E}}_{I}$ is computed by summing three types of embeddings: visual feature embedding, visual geometry embedding and modality segment embedding. We first extract the visual features $\textit{\textbf{V}}=\{\textit{\textbf{v}}_1, \textit{\textbf{v}}_2, ..., \textit{\textbf{v}}_n\}$ for all candidates using Faster R-CNN (pooled RoI features), and build a geometry feature $\textit{\textbf{G}}=\{\textit{\textbf{g}}_1, \textit{\textbf{g}}_2, ..., \textit{\textbf{g}}_n\}$ for all candidates, where $\textit{\textbf{g}}_i$ is a 7-dimensional vector consisting of the geometry information of the bounding box corresponding to candidate $c_i$, namely normalized top, left, bottom, right coordinates, width, height, and area, denoted by $\textit{\textbf{g}}_i = [x1, y1, x2, y2, w, h, w * h]$. Visual feature embeddings and visual geometry embeddings are generated by mapping the visual features and the geometry features into the same vector space through a fully connection layer $fc$:
\begin{equation}
\textit{\textbf{E}}_{I}=LN(fc(\textit{\textbf{V}})+fc(\textit{\textbf{G}})+\textit{\textbf{M}}_I)
\end{equation} where $LN$ is the layer normalization layer and $\textit{\textbf{M}}_I$ is the modality segment embedding for the image input (like segment embedding for two sentence in BERT model).

\paragraph{Text embedding} Similarly, the text embedding $\textit{\textbf{E}}_{T}$ is computed based on three different types of embeddings: token embedding, position embedding and modality embedding (Normally there is a fourth embedding, sentence segment embedding similarly to BERT, but, in our task, both expressions and facts consist of one sentence only and so only the first sentence segment embedding is used). Similar to BERT \cite{devlin2018bert}, the text $W = \{w_1, w_2, ..., w_u\}$ is first tokenized by WordPieces \cite{wu2016google}, which are then built into token embeddings $\textit{\textbf{T}} = \{\textit{\textbf{t}}_1, \textit{\textbf{t}}_2, ..., \textit{\textbf{t}}_v\}$ and position embeddings $\textit{\textbf{P}} = \{\textit{\textbf{p}}_1, \textit{\textbf{p}}_2, ..., \textit{\textbf{p}}_v\}$ according to their position in the text sequence.
\begin{equation}
\textit{\textbf{E}}_{T}=LN(\textit{\textbf{T}}+\textit{\textbf{P}}+\textit{\textbf{M}}_T)
\end{equation} where $\textit{\textbf{M}}_T$ is the modality segment embedding for the text input. 

\paragraph{Input embedding} The final input embedding $\textit{\textbf{E}}_{Inp}$ is computed by concatenating image embedding $\textit{\textbf{E}}_{I}$ and text embedding $\textit{\textbf{E}}_{T}$: 
\begin{equation}
\textit{\textbf{E}}_{Inp} = 
[\textit{\textbf{E}}_{I}, \textit{\textbf{E}}_{T}]
\end{equation}

\section{Datasets}
\label{sec:appendix Dataset}
We use the KB-Ref dataset \cite{wang2020give} aiming at evaluating the task of referring expression comprehension based on commonsense knowledge. KB-Ref consists of 43,284 expressions for 1,805 object categories on 16,917 images, as well as a knowledge base of key--value (category--fact) pairs collected from three common knowledge resources: Wikipedia, ConceptNet \cite{speer2017conceptnet} and WebChild \cite{tandon2017webchild}). KB-Ref is split into a training set (31,284 expressions with 9,925 images), a validation set (4,000 expressions with 2,290 images) and a test set (8,000 expressions with 4,702 images).

We furthermore introduce commonsense knowledge into traditional tasks/datasets of referring expression comprehension, namely RefCOCO, RefCOCOg and RefCOCO+ \footnote{following Apache License 2.0}. The datasets are devised from the MSCOCO image dataset \cite{lin2014microsoft} but differ in the types of expressions and object candidate settings. Specifically, RefCOCO+ does not allow the use of absolute location words in the expressions, and most expressions focus on the appearance of the objects. The expressions in RefCOCOg are longer and contain more 
descriptive words. RefCOCO and RefCOCO+ contain more objects of the same category within an image.

\begin{figure}
\centerline{\includegraphics[width=0.5\textwidth]{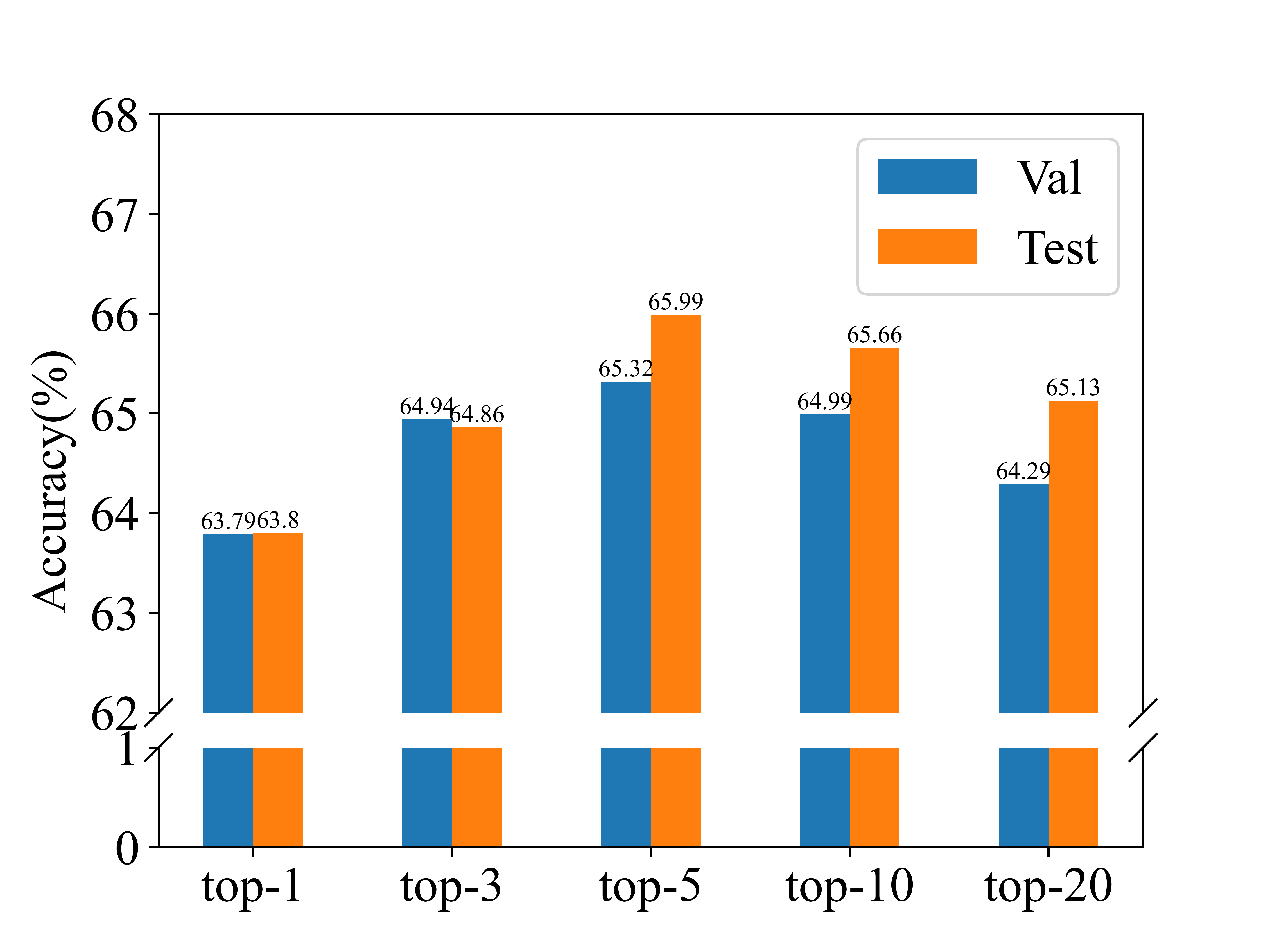}}
\caption{Accuracy across a varying number of facts (top-K). }
\label{Accuracy on different top-K facts}
\end{figure}

\section{Experimental settings}
\label{sec:appendix Experimental settings}
We extract image region features using Faster R-CNN with ResNet-101 \cite{ren2015faster}  which was pre-trained on Visual Genome \cite{krishna2016visual} using object and attribute annotations \cite{anderson2018bottom}. 
For bounding box detection, we keep the bounding boxes with at least 0.2 confidence score indicating the extent of detection. In the CK-T, the hidden layer dimension is $768$ and the number of multi-head attention heads is $12$. The models are trained using Adamw \cite{loshchilov2017decoupled} with a learning rate of $6e^{-5}$ and a batch size of $64$ on Titan RTX GPUs. Our CK-Transformer has 120M parameters in total where fact-aware classifier has 34M and bi-modal encoder has 86M. As for UNITER model, we use same setting with \textit{UNITER-base}, except for using Nvidia Apex\footnote{https://github.com/NVIDIA/apex} for speeding up training. The efficiency of our model is effected by the number of facts. Specifically, we train our CK-Transformer 10000 steps and a batch per step, which takes 2.5, 3, 3.5, 7 days with the number of facts: 3, 5, 10, 20 respectively. The CK-Transformer trains 3.8, 2.8, 2.1, 0.7 sample in average per second and tests 8.3, 7.3, 6.6, 1.1 sample per second.

\begin{figure}
\centerline{\includegraphics[width=0.5\textwidth]{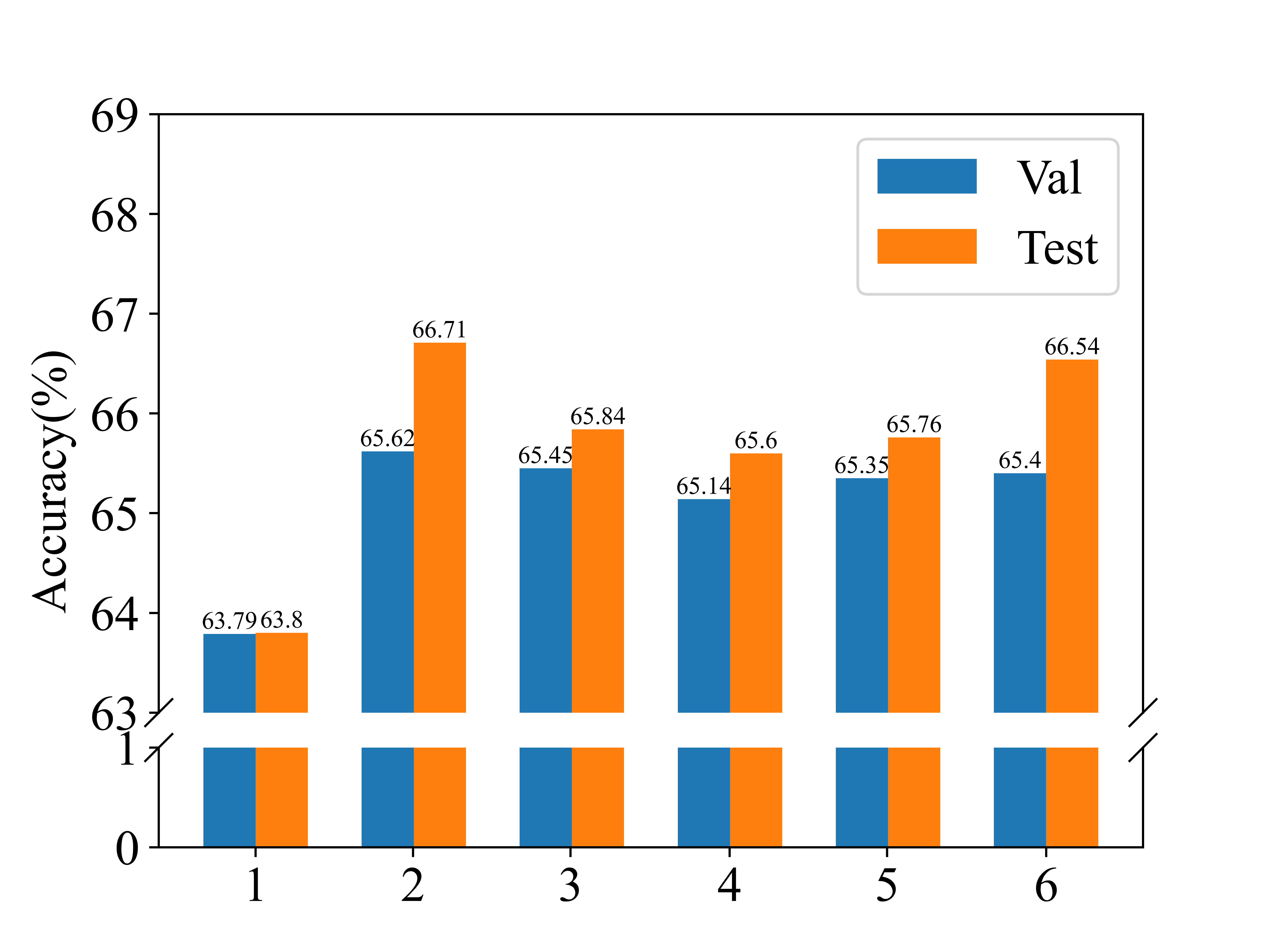}}
\caption{Accuracy across a varying number of fact-aware classifier block (M).}
\label{Accuracy on different integrator layer}
\end{figure}

\section{Impact of CK-T structure}
\label{sec:appendix parameter search}

We explore the impact in performance on KB-Ref as we vary the number of top-K facts (K) and fact-aware classifier block (M) on the development set. We first keep the number of the fact-aware classifier block constant and set it to $1$ to experiment with different values for K from $1$ to $20$. As shown in Figure \ref{Accuracy on different top-K facts}, as K increases, performance starts to improves with a peak at K=5 before starting to gradually decrease performance. 

In the second experiment, we keep K constant and set it to $3$ and explore the effect of varying values for M. We observe that the highest accuracy is achieved with with top-3 facts and 2 integrator layers as shown in Figure \ref{Accuracy on different integrator layer}.

   
   
    

\section{Introducing facts in traditional REC tasks based on detection}
\label{sec:appendix intro facts into refcoco using detected bbxes}

The results of introducing facts in traditional REC tasks based on detected bbxes and categories are shown in Table \ref{Introducing commonsense based on detection}. Compared to result based on ground-truth bbxes and categories (Table \ref{Introducing commonsense}), the improvement on models based on detection is less or even worse than the models without facts.

\begin{table}[ht]
\centering
\begin{tabular}{llcc}
\hline 
\multicolumn{2}{c}{\multirow{2}{*}{\textbf{Task}}} & \multicolumn{2}{c}{\textbf{Accuracy (\%)}}  \\
& &  $\textbf{U}_{REC}$ & \textbf{Intro Facts} \\
\hline
\multirow{3}{*}{\begin{tabular}[c]{@{}l@{}}Ref-\\ COCO\end{tabular}}  
& Val$^d$ & \textbf{81.15} & 81.06\\
& Test A$^d$ & 86.85 & \textbf{86.87}\\
& Test B$^d$ & \textbf{74.48} & 73.97\\ \hline
\multirow{3}{*}{\begin{tabular}[c]{@{}l@{}}Ref-\\ COCO+\end{tabular}} 
& Val$^d$ & \textbf{74.74} & 74.68\\
& Test A$^d$ & \textbf{81.05} & 80.70\\
& Test B$^d$ & 65.88 & \textbf{66.07}\\ \hline
\multirow{2}{*}{\begin{tabular}[c]{@{}l@{}}Ref-\\ COCOg\end{tabular}} 
& Val$^d$ & 74.49 & \textbf{74.69}\\
& Test$^d$ & \textbf{75.24} & 74.86\\ \hline
\end{tabular}
\caption{\label{Introducing commonsense based on detection} Introducing facts into RefCOCO, RefCOCO+ and RefCOCOg based on detection ($d$).}
\end{table}

\section{McNemar Test}
We also report the statistical significance for accuracy (shown in Table \ref{Introducing commonsense}) on the tasks of RefCOCOs. Specifically, we conduct the McNemar Test between models before and after introducing facts, on the test set of RefCOCO, RefCOCO+ and RefCOCOg, respectively. As shown in Table \ref{table:McNemar Test}, as for Test set on RefCOCOg and Test A on RefCOCO \textit{p}-value = 1.19$e-$08 and \textit{p}-value = 0.049 ($<$ 0.05)respectively, which means the proportion of errors is statistically significantly different after introducing facts as compared to before. However, the change in the proportion of errors after introducing facts on other tasks (Test B on RefCOCO, Test A and Test B on RefCOCO+) is not statistically significant. This is reasonable, as the error from detection will affect the fact search (we first retrieve facts using the category) and thus more error information is introduced into CK-Transformer, which make the performance worse.

\label{sec:McNemar Test}
\begin{table}
\centering
\begin{tabular}{llc}
\hline 
\multicolumn{2}{c}{\multirow{2}{*}{\textbf{Task}}} & \multicolumn{1}{c}{\textbf{McNemar Test}}  \\
& & (\textit{p}-value) \\
\hline
\multirow{2}{*}{\begin{tabular}[c]{@{}l@{}}RefCOCO\end{tabular}} 
& Test A & 0.049\\
& Test B & 0.905\\\hline
\multirow{2}{*}{\begin{tabular}[c]{@{}l@{}}RefCOCO+\end{tabular}} 
& Test A & 0.297\\
& Test B & 0.966\\\hline
\multirow{1}{*}{\begin{tabular}[c]{@{}l@{}}RefCOCOg\end{tabular}} 
& Test & 1.19$e-$08\\\hline
\end{tabular}
\caption{\label{table:McNemar Test}The McNemar Test between models before and after introducing facts on the tasks of RefCOCOs.}
\end{table}

\section{Example searched fact using different methods}
\label{sec:appendix searched fact}
\begin{figure*}[t]
\centerline{\includegraphics[width=0.95\textwidth]{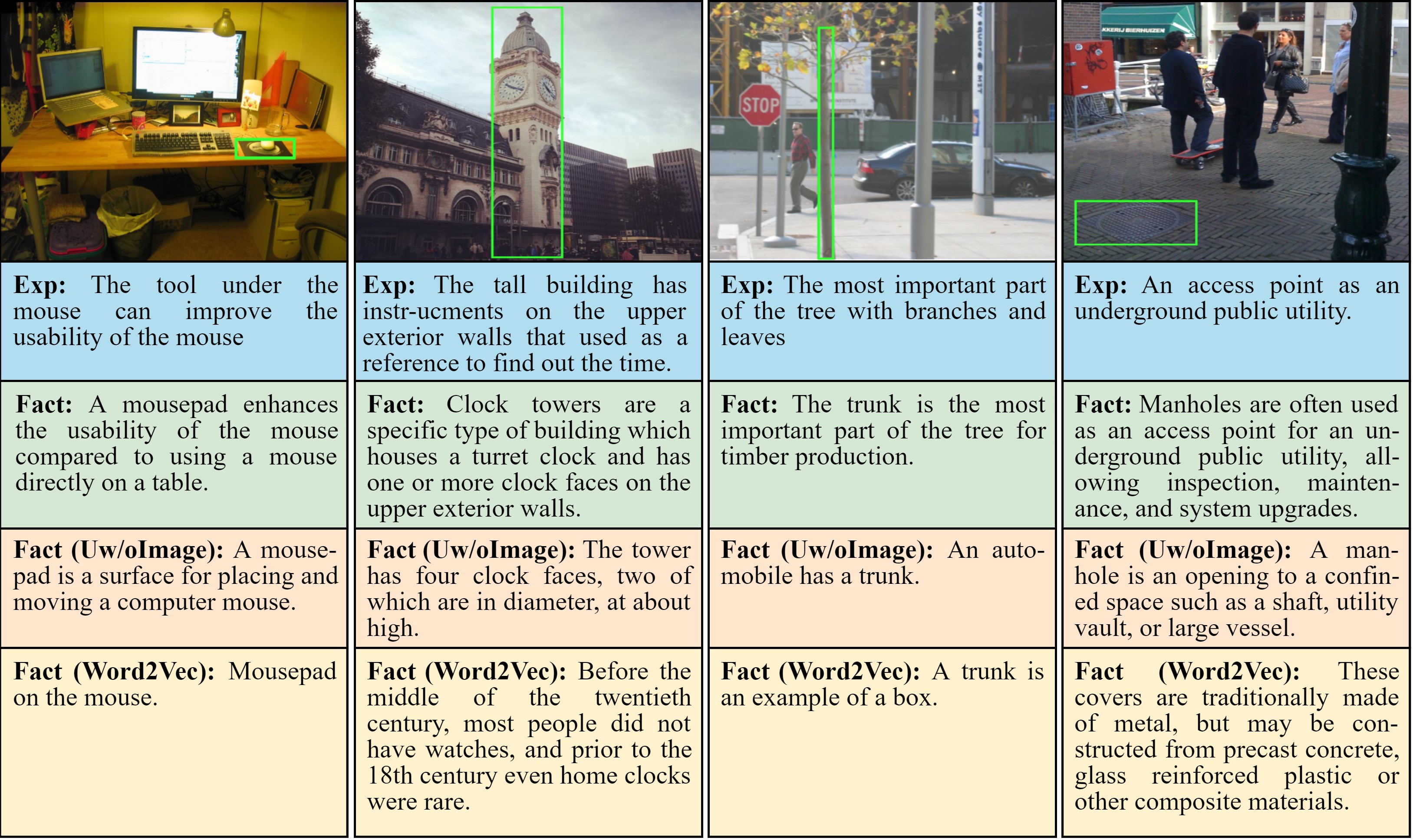}}
\caption{Example fact search process (using the top-1 fact) for different search methods: CK-T (green), CK-T-Uw/oImage (orange) and CK-T-Word2Vec (yellow).}
\label{visualizationForFacts}
\end{figure*}

As shown in Figure \ref{visualizationForFacts}, there are several facts which are selected from three different fact search methods: CK-Transformer, CK-T-Uw/oImage and CK-T-Word2Vec. As we can see in the Table, normally the facts of CK-Transformer model (green) is the best relevant with the referring expression (blue) and the facts in CK-T-Word2Vec model is the worst relevant with the expression.

\clearpage

\end{document}